\documentclass[conference]{IEEEtran}
\IEEEoverridecommandlockouts
\usepackage{cite}
\usepackage{amsmath,amssymb,amsfonts}
\usepackage{algorithmic}
\usepackage{graphicx}
\usepackage{textcomp}
\usepackage[dvipsnames]{xcolor}
\usepackage[colorlinks = true,
            linkcolor = MidnightBlue,
            urlcolor  = MidnightBlue,
            citecolor = MidnightBlue,
            anchorcolor = MidnightBlue]{hyperref}

\usepackage{balance}
\usepackage{booktabs}
\def\BibTeX{{\rm B\kern-.05em{\sc i\kern-.025em b}\kern-.08em
    T\kern-.1667em\lower.7ex\hbox{E}\kern-.125emX}}

\begin{document}
\title{\LARGE \bf Toward Aligning Human and Robot Actions via Multi-Modal Demonstration Learning
\thanks{$^1$ Department of Electrical Engineering and Computer Science, University of Tennessee Knoxville, Knoxville, TN 37996, USA {\tt\footnotesize \{azahid, fwang31, ady\}@vols.utk.edu, \{fliu33, sai\}@utk.edu}}%
\thanks{$^2$ Istituto Italiano di Tecnologia, Italy {\tt\footnotesize jie.fan.0414@gmail.com}}
}

\author{Azizul Zahid$^{1}$, Jie Fan$^{2}$, Farong Wang$^{1}$, Ashton Dy$^{1}$, Sai Swaminathan$^{1}$, Fei Liu$^{1}$
\vspace{-2mm}
}

\maketitle

\begin{abstract}
Understanding action correspondence between humans and robots is essential for evaluating alignment in decision-making, particularly in human-robot collaboration and imitation learning within unstructured environments. We propose a multimodal demonstration learning framework that explicitly models human demonstrations from RGB video with robot demonstrations in voxelized RGB-D space. Focusing on the “pick and place” task from the RH20T dataset, we utilize data from 5 users across 10 diverse scenes. Our approach combines ResNet-based visual encoding for human intention modeling and a Perceiver Transformer for voxel-based robot action prediction. After 2000 training epochs, the human model reaches 71.67\% accuracy, and the robot model achieves 71.8\% accuracy, demonstrating the framework's potential for aligning complex, multimodal human and robot behaviors in manipulation tasks. \textbf{Code available at:} \href{https://github.com/utkauraslab/aligning_hr_actions}{\texttt{github.com/utkauraslab/aligning\_hr\_actions}}


\end{abstract}


\section{Introduction}

Robots that learn manipulation skills by observing humans have the potential to eliminate labor-intensive manual programming in real-world settings. Central to achieving this vision is effectively mapping human behaviors to robotic actions—translating human demonstrations into executable robot policies \cite{Yuzhe_2022_ECCV}. Current state-of-the-art methods face two primary challenges. First, modality mismatches between human demonstrations (typically captured as 2D videos or images) and robot perception (usually in 3D via point clouds or RGB-D sensors) limit the accuracy with which robots understand spatial context for precise manipulation \cite{wang2023mimicplay, zeng2020transporter}. Second, many learning-from-demonstration (LfD) approaches depend heavily on task-specific demonstrations and manual action remapping, which restricts their scalability to diverse skill sets \cite{Lin_Shao_2021, Mandlekar_2018}.

Beyond these issues, an essential challenge underlying these limitations is the lack of clear alignment—that is, a precise correspondence between what humans demonstrate and what robots execute. Without this alignment, robots cannot reliably interpret and replicate human intent, significantly reducing the effectiveness of imitation learning. Despite its importance, this issue of alignment remains relatively unexplored, especially regarding critical aspects like accurately capturing movement trajectories, identifying reusable action components (``action primitives"), and recognizing shifts in human intention (``intention switching").


Motivated by these challenges, we introduce a new framework to map human demonstrations directly to robot actions in manipulation tasks such as pick-and-place. In our approach, we combine data from RGB video (to capture human intentions) and 3D sensor inputs (to represent the robot’s workspace) into one integrated pipeline. This end-to-end framework ensures translating human actions to robot behaviors reliably and accurately.

Our method predicts human intentions from RGB videos and robot actions from voxelized RGB-D inputs for the same high-level task, establishing a unified modeling foundation for cross-modal intention-action mapping. Specifically, we employ a two-stage model: a ResNet+LSTM network to capture the temporal evolution of human intentions from video data, and a voxel-based Perceiver Transformer \cite{shridhar2022_peract} to predict appropriate robot actions within the 3D workspace. This design bridges the gap between the 2D information in human demonstrations and the 3D perception of the robot, ensuring that high-level intentions are accurately represented across both modalities.

We evaluate our framework on a challenging eight-class object-picking task. The experimental results show robust cross-modal performance and improved semantic consistency between human demonstrations and robot actions. Our contributions are summarized as follows: 

\begin{itemize} 
\item We introduce a novel framework that directly aligns human demonstrations with robot actions using both video and RGB-D data with probabilities. 
\item We develop a two-stage model that employs a ResNet+LSTM network for extracting human intention sequences and a voxel-based Perceiver Transformer for robot action prediction.
\item We empirically evaluate our framework on an eight-class object-picking task, showing consistent performance across modalities.
\end{itemize}
Our work showcase the potential of aligning human and robot behaviors across modalities (e.g., 2D + RGB-D data) as a foundation for advancing imitation learning in robotics.

\begin{figure*}[!htb]
    \centering
    \includegraphics[width=0.95\linewidth]{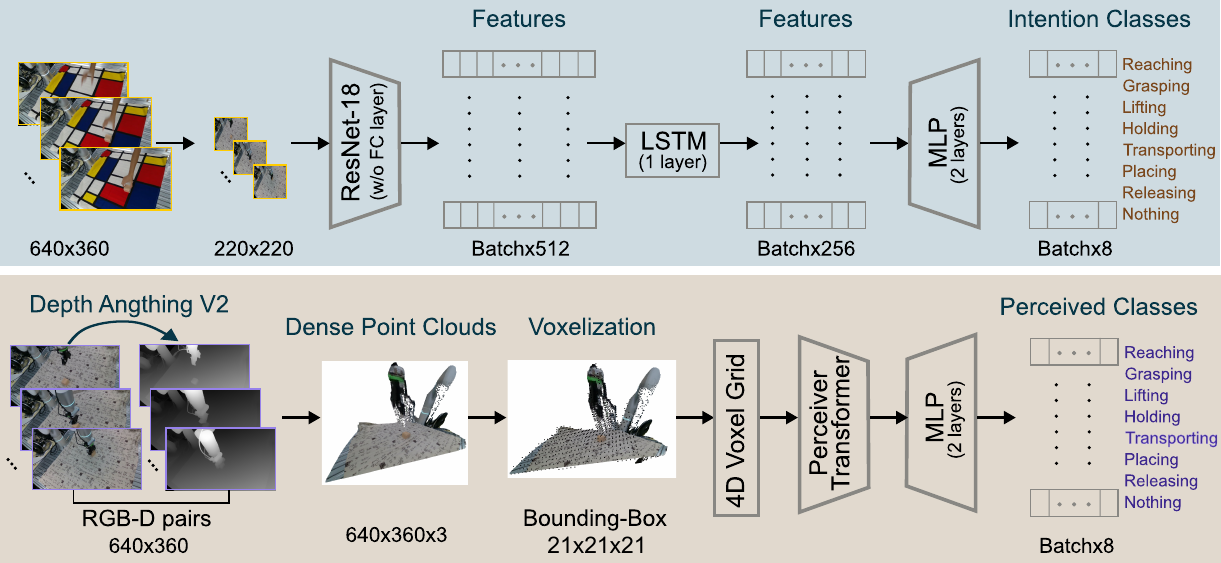} 
    \caption{Human and Robot demonstration framework.}
    \label{fig:framework}
    \vspace{-2mm}
\end{figure*}
\section{Methodology}

\subsection{Problem Formulation}
We aim to learn a mapping between human and robot behaviors. Let human demonstrations be represented as a sequence of RGB frames $H = \{h_t\}_{t=1}^T$, and robot demonstrations as voxelized RGB-D inputs $R = \{r_t\}_{t=1}^T$. We define a set of human intention classes $\{i_1, i_2, ..., i_N\}$ and robot action classes $\{j_1, j_2, ..., j_N\}$, and aiming for their semantic correspondence for the future.

For this study, robot and human share the same set of intention-action classes. Our focus is on a high-level manipulation task, i.e., ``pick" task, for which we consider eight intention classes ($N=8$): \textit{Reaching}, \textit{Grasping}, \textit{Lifting}, \textit{Holding}, \textit{Transporting}, \textit{Placing}, \textit{Releasing}, and \textit{Nothing}.

\subsection{Human Demonstration Encoding}

We extract RGB frames \( h_t \) from human demonstration videos (originally 640\(\times\)360 pixels), resized and normalized to meet the input requirements ($220 \times 220$ pixels) of the ResNet-18 model. These frames are passed through a pretrained ResNet-18 \cite{he2016deep}, without the final fully connected layers, to obtain feature representations of size \( (b, 512) \), where \( b \) denotes the batch size.

To capture temporal dynamics across frames, we utilize a single-layer LSTM network. This LSTM encodes sequential dynamics inherent to the video data, allowing the model to discern subtle temporal patterns characteristic of object manipulation sequences, i.e., grasping, lifting, transporting, and placing. It outputs feature representations of dimension $(b, 256)$, effectively embedding enriched temporal information.

These encoded LSTM features $(\mathcal{Z}^{H}_t)$ are then passed to a softmax-based multilayer perceptron (MLP) classifier for intention recognition. The MLP consists of two hidden layers, transforming the input from \( (b, 256) \) to \( (b, 128) \), and finally to \( (b, 8) \), corresponding to the eight intention classes. The $P(i_t = c | H)$  is the probability of human intention at time t, where $c$ is the predefined 8 intention-action classes.

\vspace{-1 em}

\begin{equation}
\begin{split}
    \mathcal{F}^H_t & = f_{\text{CNN}}(h_t) \\
    \quad \mathcal{Z}^H_t & = f_{\text{LSTM}}(\mathcal{F}^H_{1:t}) \\
    \mathbf{P}(i_t = c | H) & = \frac{\exp(\text{MLP}_H(\mathcal{Z}^H_t)[c])}{\sum_{c'} \exp(\text{MLP}_H(\mathcal{Z}^H_t)[c'])}
\end{split}
\end{equation}

\vspace{-1 em}



\subsection{Robot Demonstration Encoding}
We 
generate depth images from RGB inputs using Depth Anything V2 \cite{depth_anything_v2}, a transformer-based monocular depth estimation model. These depth images are aligned with their respective RGB frames to produce RGB-D datasets.

From each RGB-D pair, we create dense point clouds utilizing intrinsic camera parameters and following the Open3D back-projection model. Each point cloud is subsequently projected onto the original image plane to retrieve corresponding RGB values, which are then assigned as color features to each point.

To construct a structured and consistent representation from the unstructured point cloud data, we apply voxelization. Using a bounding-box-based method (box size: $100 \times 100 \times 100$), points are first normalized within a predefined spatial region and then mapped to voxel cells according to their spatial indices. RGB values act as feature vectors for each point. Within each voxel, points are aggregated through mean pooling, yielding a feature vector of dimension $C$ per voxel.

Formally, given a point cloud with $N$ points and features $x_n \in \mathbb{R}^C$, and bounding volume $[x_{\min}, x_{\max}] \times [y_{\min}, y_{\max}] \times [z_{\min}, z_{\max}]$, we define a fixed resolution voxel grid of shape $(D, H, W)$. Each voxel is populated as:
\vspace{-0.5 em}
\begin{equation}
    V_{ijk} = \frac{1}{|S_{ijk}|} \sum_{x_n \in S_{ijk}} x_n
\end{equation}
\vspace{-0.2 em}
where $S_{ijk}$ is the set of points falling into voxel cell $(i,j,k)$. An additional occupancy channel is appended to each voxel to indicate whether the voxel is non-empty.

The resulting 4D voxel grid $V_t \in \mathbb{R}^{B \times H \times W \times C}$ is then flattened and passed to a Perceiver Transformer \cite{shridhar2022_peract}:
\vspace{-0.2 em}
\begin{equation}
    \mathcal{Z}^R_t = f_{\text{Perceiver}}(V_t)
\end{equation}
A softmax classifier then predicts robot actions:
\begin{equation}
    \mathbf{P}(j_t = c | R) = \frac{\exp(\text{MLP}_R(\mathcal{Z}^R_t)[c])}{\sum_{c'} \exp(\text{MLP}_R(\mathcal{Z}^R_t)[c'])}
\end{equation}
\vspace{-0.5 em}




\section{Experiments and Results}
We evaluate our model on the ``pick" task using the RH20T dataset, which includes RGB human demonstrations and RGB-D robot observations from 5 users across 10 scenes,
both labeled with one of eight intention/action classes.
Both human and robot branches are trained independently. We use cross-entropy loss for both classification heads, with class weights to account for class imbalance. The human branch uses ResNet-18 feature extraction followed by an LSTM, while the robot branch uses voxelized 3D observations processed with a Perceiver model.

\subsection{Human Intention Model}

\subsubsection{Training Details}
\label{subsubsec:training}

We collected 70 demonstration videos from 5 users, with each user contributing up to 10 scenes for different ``pick'' tasks. Each video varies in duration, resulting in a different number of frames per sequence. After extracting frames from each video, we treated them as 70 separate sequences. Each frame was resized$(220, 220)$, normalized, and converted to a tensor to be compatible with the ResNet-18 architecture.

We used a pretrained ResNet-18 model (with the final fully connected layers removed) to extract a 512-dimensional feature vector for each frame. These frame-wise features were then passed through our LSTM+MLP model to perform intention classification. 

\begin{figure}[t]
    \centering
    \includegraphics[width=0.99\linewidth]{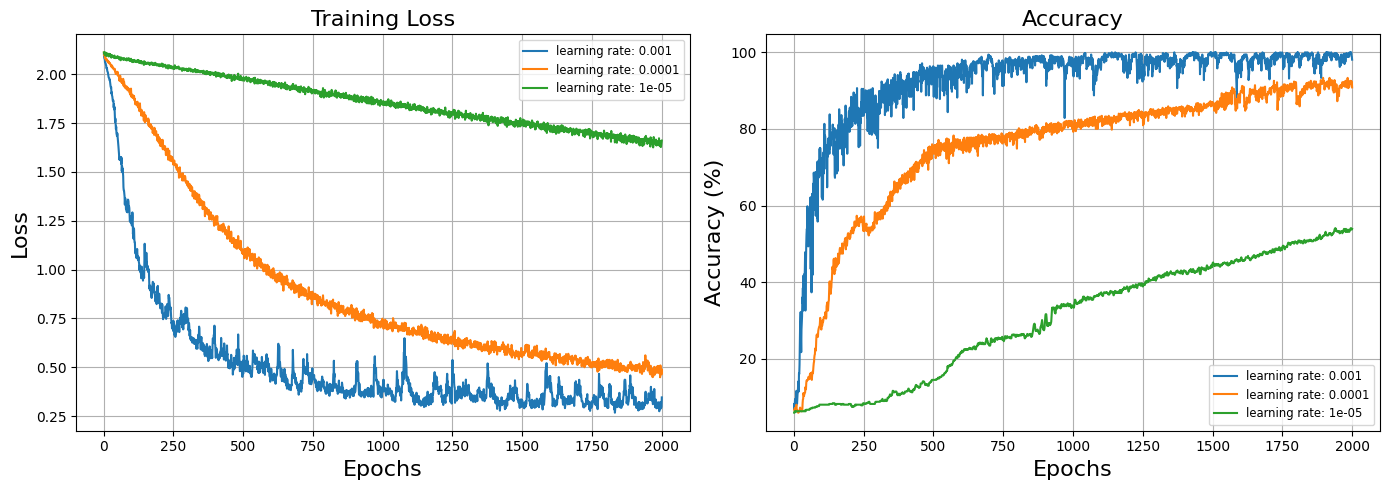}
    \caption{Training loss and validation accuracy trend for three different learning rates: 0.001(blue), 0.0001(orange), 0.00001(green).}
    \label{fig:learning_rate}
    \vspace{-5mm}
\end{figure}

We conducted an extensive hyperparameter search (Table \ref{tab:hyperparameters_human}), varying the LSTM hidden state size, number of LSTM layers, batch size, and learning rate. The dataset was split into 70\% for training, 20\% for validation, and 10\% for testing. The highest validation accuracy (71.67\%) was achieved with the configuration: hidden size 64, learning rate 0.001, batch size 32, and LSTM layer 1. However, this setting exhibited significant instability and loss fluctuations during training, as shown in Figure \ref{fig:learning_rate}. Despite its better convergence, the frequent loss spikes made it less reliable. In contrast, the second-best configuration—hidden size 64, learning rate 0.0001, batch size 16, and 1 LSTM layer—achieved a more stable performance with a test accuracy of 63.95\%, and was therefore selected as our final model.

Other hyperparameters, such as the number of training epochs (2000), Adam optimizer, and a dropout rate of 0.7, were kept fixed throughout the experimentation.


\begin{table}[h]
\centering
\caption{Hyperparameter configurations and their corresponding accuracy for the Human Intention Model.}
\label{tab:hyperparameters_human}
\begin{tabular}{p{1.2cm}p{1.5cm}p{1.2cm}p{1.3cm}p{1.2cm}}
\toprule
\textbf{Hidden Size} & \textbf{Learning Rate} & \textbf{Batch Size} & \textbf{LSTM Layers} & \textbf{Accuracy (\%)} \\
\midrule
256 & 0.0001 & 32 & 1 & 58.37 \\
128 & 0.0001 & 32 & 1 & 61.37 \\
\textbf{64}  & 0.0001 & 32 & 1 & \textbf{63.09} \\
\midrule
64  & \textbf{0.001} & 32 & 1 & \textbf{71.67} \\
64  & 0.0001 & 32 & 1 & 63.09 \\
64  & 0.00001 & 32 & 1 & 37.34 \\
\midrule
64  & 0.0001 & \textbf{16} & 1 & \textbf{63.95} \\
64  & 0.0001 & 32 & 1 & 63.09 \\
64  & 0.0001 & 64 & 1 & 62.66 \\
\midrule
64  & 0.0001 & 16 & \textbf{1} & \textbf{63.95} \\
64  & 0.0001 & 16 & 2 & 62.23 \\
64  & 0.0001 & 16 & 3 & 53.65 \\
\bottomrule
\end{tabular}
\end{table}

\subsubsection{Results}
We evaluated the performance of our human intention model in intention classification using per-class accuracy and a confusion matrix. The model was trained with various hyperparameter configurations, as detailed in Section \ref{subsubsec:training}.


\begin{figure}[!htb]
    \centering
    \includegraphics[width=1.0\linewidth]{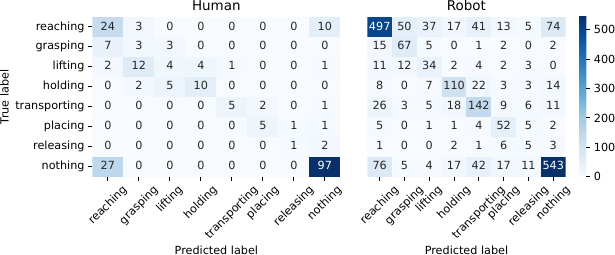}
    \vspace{-5mm}
    \caption{Confusion matrices for 8 intention-action classes of human/robot data.}
    \label{fig:confusion_human_robot}
\end{figure}
Accuracies range from 16.67 to 78.23 (in Table \ref{tab:accuracy_comparison}, with nothing achieving the highest (78.23), likely due to class prevalence and distinct features. reaching (64.86) and transporting (62.50) also show strong performance, suggesting clear temporal patterns. In contrast, the model struggles with grasping (23.08) and lifting (16.67), likely due to overlapping dynamics, class imbalance, and the fact that grasping and lifting have the lowest amount of data in the training set. Additionally, grasping is often confused with reaching, as these two are sequential intentions, and their labelings are temporally very close most of the time, as evident from the confusion matrix (in Figure \ref{fig:confusion_human_robot}) where 7 out of 13 grasping instances are misclassified as reaching. Lifting also faces similar issues, with 12 out of 24 instances misclassified as grasping, reflecting their sequential nature and temporal proximity.

\subsection{Robot Action Model}

\subsubsection{Training Details}

To classify robot actions from voxelized RGB-D observations, we convert each RGB-depth pair into a 3D voxel grid using a bounding-box-based voxelization scheme. Each voxel encodes the mean RGB of contained points, with an added occupancy channel indicating whether it is occupied. The resulting 4D voxel grids ($D \times H \times W \times C$) are flattened into sequences and processed by a Perceiver Transformer, which uses cross-attention to project inputs into a latent space. Latent features are passed through a multi-layer perceptron (MLP) for final classification.

The dataset includes segmented robot actions labeled with eight action classes as the human demonstration. Data is split into 70\% training, 20\% validation, and 10\% testing. The model is trained for 150 epochs using the Adam optimizer (learning rate $1\mathrm{e}{-4}$, batch size 10) on an NVIDIA RTX 4090 GPU.

\subsubsection{Hyperparameter Tuning}
We adopted a fixed model configuration after preliminary tuning, constrained by hardware limits: input dimension 10 (voxel feature size), latent dimension 512, 128 latent tokens, 8 cross-attention heads, and a maximum of 9,261 points per sample. To mitigate class imbalance, we apply inverse-frequency class weights in the cross-entropy loss.

\subsubsection{Results}
\begin{figure}[t]
    \centering
    \includegraphics[width=0.85\linewidth]{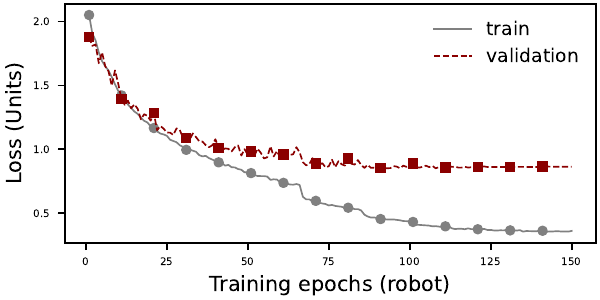}
    \caption{Training and validation loss curves on voxelized robot action data.}
    \label{fig:robot_loss}
    \vspace{-5mm}
\end{figure}

The model converged stably over 150 epochs (Figure~\ref{fig:robot_loss}), reaching an average validation accuracy of \textbf{71.80\%}, comparable to the human prediction branch. Training and validation losses dropped rapidly during early epochs, indicating fast convergence. After epoch 100, training loss continued to decrease and stabilized near zero. Validation loss, though more variable due to small batch size and class imbalance, followed a similar downward trend with occasional spikes—likely from underrepresented or ambiguous classes. The close alignment between training and validation curves suggests strong generalization with minimal overfitting.

\begin{figure}[!htb]
    \centering
    \includegraphics[width=0.99\linewidth]{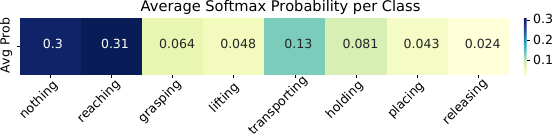}
    \vspace{-5mm}
    \caption{Predicted softmax probability for each robot action class.}
    \label{fig:robot_probs}
\end{figure}
To assess prediction confidence, we visualize softmax outputs for selected validation samples in Figure~\ref{fig:robot_probs}. The model exhibits high confidence for classes like \textit{nothing} and \textit{holding}, while early-stage actions such as \textit{grasping} and \textit{lifting}, along with late-stage actions like \textit{placing} and \textit{releasing}, show lower confidence and increased ambiguity. This may be attributed to subtle inter-class motion differences and visual similarities among transitional actions, which make them harder to distinguish based solely on voxelized spatial cues at individual time steps. In contrast, static or sustained postures, such as \textit{reaching} or \textit{nothing}, tend to produce more consistent geometric patterns, leading to more confident predictions.

\subsubsection{Per-Class Accuracy Analysis}

\begin{table}[!htb]
\centering
\caption{Prediction accuracy comparison across 8 classes for human intentions and robot actions (10 samples per class).}
\label{tab:accuracy_comparison}
\begin{tabular}{lcc}
\toprule
\textbf{Class} & \textbf{Human Accuracy (\%)} & \textbf{Robot Accuracy (\%)} \\
\midrule
Reaching     & 64.86 & 50.0 \\
Grasping     & 23.08 & 83.0 \\
Lifting      & 16.67 & 100.0 \\
Holding      & 58.82 & 100.0 \\
Transporting & 62.50 & 67.0 \\
Placing      & 71.43 & 56.0 \\
Releasing    & 33.33 & 75.0 \\
Nothing      & 78.23 & 80.0 \\
\midrule
\textbf{Average} & \textbf{63.95} & \textbf{76.00} \\
\bottomrule
\end{tabular}
\end{table}

Table~\ref{tab:accuracy_comparison} presents the per-class prediction accuracy for both human intention recognition and robot action classification across eight semantic categories. Each class contains 10 validation cases.

The robot branch consistently outperforms the human branch, achieving an average accuracy of \textbf{76.00\%} compared to \textbf{63.95\%} for human predictions. Notably, the robot model achieves perfect accuracy on \textit{lifting} and \textit{holding}, and high accuracy on \textit{grasping} (83.0\%) and \textit{releasing} (75.0\%). These results suggest that voxelized RGB-D representations provide strong spatial cues for action discrimination, particularly for mid- and late-stage manipulation steps.

In contrast, the human model struggles with early-stage intentions such as \textit{grasping} (23.08\%) and \textit{lifting} (16.67\%), likely due to the visual ambiguity and subtle motion cues present in RGB-only videos. The highest performance in the human branch is observed for \textit{nothing} (78.23\%) and \textit{placing} (71.43\%), which may exhibit more distinctive visual patterns or static poses.

Overall, these results highlight the strength of the voxel-based robot encoder in capturing geometric and depth-aware action features, while also pointing to the need for improved temporal modeling and motion-sensitive features in the human branch, particularly for fine-grained early-stage actions.

\section{Discussion and Future Work}

For the next step, we aim to validate semantic correspondence learning by jointly modeling human intentions and robot actions. To quantify their alignment, we define an alignment score \(S(H, R)\) that aggregates the joint confidence of corresponding predictions across a temporal sequence:
\begin{equation}
    S(H, R) = \frac{1}{T} \sum_{t=1}^{T} \delta(i_t, j_t)\, \mathbf{P}(i_t | H)\, \mathbf{P}(j_t | R),
\end{equation}
where \(\delta(i_t, j_t)\) is an indicator function denoting semantic consistency between the predicted human intention \(i_t\) and robot action \(j_t\). The score encourages high confidence for semantically aligned predictions over time. We propose to jointly train the human and robot models by maximizing this alignment score with respect to model parameters \(\theta\):
\[
\theta^* = \arg\max_{\theta} S(H, R; \theta).
\]
This framework lays the groundwork for learning cross-modal behavioral correspondence and paves the way for joint action reasoning in multimodal human-robot learning.

Our initial results reveal both strengths and limitations of the proposed intention-action models. The human intention model performs well on visually distinct or frequent classes like \textit{nothing}, \textit{holding}, and \textit{transporting}, but struggles with \textit{grasping}/\textit{lifting} and \textit{placing}/\textit{releasing} due to temporal similarity and data imbalance. This suggests current visual features may not capture the subtle cues needed for early- and late-stage action discrimination. On the robot side, the Perceiver Transformer effectively classifies voxelized RGB-D inputs, but its lack of temporal modeling limits performance on fine-grained transitions (e.g., \textit{transporting} vs. \textit{placing}), and voxel sparsity may cause information loss. To address these issues, we plan to incorporate temporal voxel sequences, point-based attention, and integrate voxel-level features togther with human motion embeddings to strengthen cross-modal alignment.

Ultimately, we aim to build a unified multimodal framework that aligns human intentions and robot actions using RGB and RGB-D inputs. Combining motion encoders such as FlowNet \cite{IMKDB17}, we seek to enable robust imitation learning from human demonstration for robotic manipulation.

\bibliographystyle{IEEEtran}
\bibliography{reference}

\end{document}